# Explainable AI in Grassland Monitoring: Enhancing Model Performance and Domain Adaptability

Shanghua Liu[1], Anna Hedström[1,2,3], Deepak Hanike Basavegowda[1,2] , Cornelia Weltzien[1,2], Marina M.-C. Höhne[1,3,4]

**Abstract:** Grasslands are known for their high biodiversity and ability to provide multiple ecosystem services. Challenges in automating the identification of indicator plants are key obstacles to large-scale grassland monitoring. These challenges stem from the scarcity of extensive datasets, the distributional shifts between generic and grassland-specific datasets, and the inherent opacity of deep learning models. This paper delves into the latter two challenges, with a specific focus on transfer learning and eXplainable Artificial Intelligence (XAI) approaches to grassland monitoring, highlighting the novelty of XAI in this domain. We analyze various transfer learning methods to bridge the distributional gaps between generic and grassland-specific datasets. Additionally, we showcase how explainable AI techniques can unveil the model's domain adaptation capabilities, employing quantitative assessments to evaluate the model's proficiency in accurately centering relevant input features around the object of interest. This research contributes valuable insights for enhancing model performance through transfer learning and measuring domain adaptability with explainable AI, showing significant promise for broader applications within the agricultural community.

## 1   Introduction

Grasslands, covering 34% of the agricultural landscape in Europe [Eu20], have recently gained significant attention due to their vital role in conserving local biodiversity, ensuring food production, and impacting ecological processes like water and climate regulation [Be19]. However, these ecosystems have been steadily declining due to intensive agricultural practices, high livestock density, abandonment, and afforestation [Sc22].

The new Green Architecture, integrated into the Common Agricultural Policy (CAP) of the European Union, places increased emphasis on halting the loss of farmland biodiversity through the protection and restoration of landscape features and semi-natural areas, including grasslands [Pe21]. For example, as part of the Green Architecture,

[1] Leibniz-Institut für Agrartechnik und Bioökonomie e.V. (ATB), Max-Eyth-Allee 100, 14469 Potsdam, Germany {SLiu; AHedstroem; DBasavegowda; CWeltzien; MHoehne}@atb-potsdam.de,
https://orcid.org/0000-0003-1692-2256
[2] Technische Universität Berlin, Straße des 17. Juni 135, 10623 Berlin, Germany
[3] Understandable Machine Intelligence Lab (UMI Lab)
[4] Universität Potsdam, An der Bahn 2, 14476 Potsdam, Germany

Germany has introduced Eco-Scheme 5 to preserve biodiversity in permanent grasslands. The scheme is a result-based approach to support farmers for their contribution to promoting biodiversity in grasslands. The only decisive factor for the payment of financial incentives is the presence of at least four indicator species typical for the region [Eu23]. While this scheme provides a strong motivation for farmers to improve grassland biodiversity, it also comes with a significant monitoring cost, as manual assessment of indicator species consumes both time and resources. Hence, it is crucial to develop efficient and reliable detection methods for these indicator species, to facilitate broader applications in grassland monitoring.

However, the development of indicator species detection methods poses significant challenges. First, the scarcity of large-scale grassland datasets arises from the labor-intensive and time-consuming nature of data collection and annotation, particularly in wild grassland settings. Second, due to the data distributional shifts, deep learning models initially trained on generic datasets like COCO should be adapted with agricultural data to match the grassland context. Third, the black-box nature of deep learning models impedes the ability of researchers to understand their predictive behaviors, limiting their practicality. Against this backdrop of challenges, our contributions are as follows:

- First, we present an indicator species detection method that harnesses the power of EfficientDet [TPL20] and leverages a grassland dataset proposed by Basavegowda et al. [Ba22], including the data from greenhouse, an experimental grassland, and semi-natural grasslands. To tackle the data distributional shift between generic datasets and grassland scenarios, we investigate the impact of different transfer learning methods on the model performance.

- Second, we reveal the black box characteristic of the Deep Learning object detection models by applying cutting-edge explainable AI techniques [He23a] [Se17] [He23b] to the grassland domain. Furthermore, by employing localization metrics [Ko20] [TMBE22], we provide novel insights into model behavior and assess its adaptability to other domains, such as grassland habitats.

## 2    Preliminaries

In this section, we explore the practical application of advanced AI technologies in agriculture. We highlight the utilization of EfficientDet, an object detection model, and showcase its effectiveness across various agricultural applications in Section 2.1. In addition, in Section 2.2, we discuss how transfer learning contributes to developing applications in the agricultural field, leveraging knowledge from a source domain to enhance learning in the agricultural domain (target domain). Furthermore, we address the significance of explainable artificial intelligence (XAI) in ensuring the transparency and interpretability of model behavior in Section 2.3.

### 2.1 EfficientDet

EfficientDet models are a family of versatile and efficient object detectors with network parameters ranging from 3.9M to 52M [TPL20]. As illustrated in Figure 1, EfficientDet comprises three integral architecture components: (i) a feature extraction network that takes image inputs and produces a series of feature maps derived from various network depths, (ii) a weighted Bi-directional Feature Pyramid Network (BiFPN), which combines multi-scale feature maps from the preceding network, assigning them weights based on the learned importance, and (iii) a detection head network that employs the input fused feature maps for object detection.

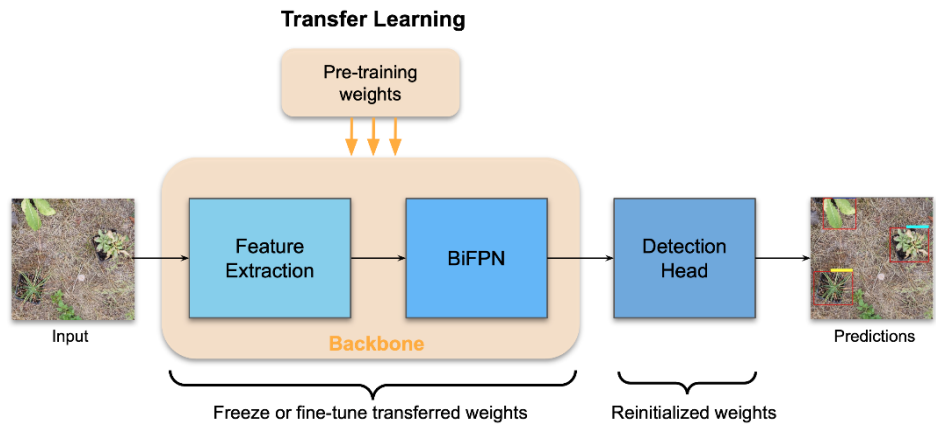

Fig. 1: EfficientDet Architecture (blue) & Transfer Learning Pipeline (orange). *From left to right*: EfficientDet consists of Feature Extraction Net, BiFPN, and Detection Head Net. *From top to bottom*: Illustration of the transfer learning process where the backbone is initialized with pre-training weights and the detection head is randomly initialized. The transferred weights can be frozen or further fine-tuned with the target dataset.

**Related Works.** EfficientDet has seen extensive applications in the field of agriculture, spanning tasks such as crop monitoring [Ba21] [Ba22] [Ca20], plant disease detection [RS21], and pest management [Č23] [AU22]. Specifically, Badeka et al. [Ba21] applied EfficientDet-D0 to detect vineyard trunks and the experimental results revealed its superiority over other detection methods like MobileNets [Ho17] and YOLOv5 [Jo21] in terms of both accuracy and running time. For disease detection tasks, R et al. [RS21] conducted a comprehensive study using three different benchmark object detection models, including EfficientDet-D2 to automatically detect leaf diseases in apples and grapes. Čirjak et al. [Č23] harnessed EfficientDet-D4 to detect pest populations in apple production for early damage detection. The wide-ranging applications of EfficientDet in agriculture serve as compelling motivation for our own utilization of EfficientDet for indicator species detection.

**2.2     Transfer Learning**

In real-world scenarios, amassing substantial datasets can be challenging due to the labor-intensive and time-consuming nature of dataset preparation. Here transfer learning enables the use of large deep learning models that were trained on a source domain with extensive dataset, such as ImageNet [De09], comprising 1.2 million images across 1000 categories, and adapted to a target domain with no or limited amount of labeled data. Hence the underlying principle is to enhance learning within the target domain by leveraging knowledge from a source domain. Figure 1 illustrates the pre-training and fine-tuning pipeline in the context of transfer learning. In this process, the weights of the backbone layer are initialized with pre-training weights learned from a substantial amount of data in the source domain. In contrast, the remaining layers are initialized randomly. Subsequently, these randomly initialized layers are trained on a relatively smaller target dataset, with the flexibility to freeze or fine-tune the parameters of the backbone layer. Freezing the parameters entails maintaining the pre-training weights unchanged, without updates during the backpropagation process of training. The pretrained weights used in our work were initially trained on the COCO dataset.

**Related Works.** Sahili et al. [ASA22] conducted an insightful investigation into transfer learning in the agricultural domain, introducing a plant-domain-specific dataset, AgriNet. They fine-tuned deep learning models such as Inception [Sz16] and Xception [Ch17], which were initially pre-trained on the ImageNet (source domain), using the AgriNet dataset (target domain). This approach involved freezing the first several layers as fixed feature extractors and randomly initializing the remaining layers. The study demonstrated the substantial performance improvements attainable through transfer learning when classifying 423 plant species, diseases, pests, and weeds. Given the demonstrated potential of transfer learning, our research seeks to investigate how different transfer learning strategies impact model performance in the context of grassland scenarios.

**2.3     eXplainable Artificial Intelligence (XAI)**

In agriculture applications such as cereal plant head detection [Sa23], citrus pest detection [Qi23], and leaf disease recognition [TLN23], the use of artificial intelligence has witnessed a remarkable increase over the past few years. Nevertheless, the pervasive use of deep learning models often presents a challenge: these models operate as enigmatic black boxes, concealing the rationale behind their predictions. In certain contexts, the ability to provide interpretable predictions holds far more significance than the relentless pursuit of accuracy through opaque models [Ru22]. This is where XAI emerges as a pivotal player, endeavoring to augment the interpretability of intricate models while retaining the fidelity of their predictions [CPC19].

**Related Works.** Recent advancements in this research domain are exemplified by the works of Wei et al. [We22] and Akagi et al. [Ak20]. Wei harnessed deep learning models to classify leaf diseases, and Akagi focused on the classification of calyx-end cracking in persimmon fruits. Both researchers integrated XAI tools to probe the depths of the models' feature extraction capabilities. However, it is noteworthy that their analysis of attention distribution within inputs remained predominantly qualitative, relying on visualization, without a quantitative evaluation. Quantus [He23a] offers a comprehensive XAI evaluation framework, equipped with quantitative evaluation metrics that provide profound insights into model behavior, delivered in an easy-to-use API. In this paper, beyond offering qualitative visualizations, we take a step forward and pioneer Quantus to species identification. We mainly focus on the localization metrics, namely *Attribution Localization* [Ko20] and *Top-K Intersection* [TMBE22], which provide valuable insights into the model's adaptability to other domains. The exploration of additional metrics is reserved for future research work.

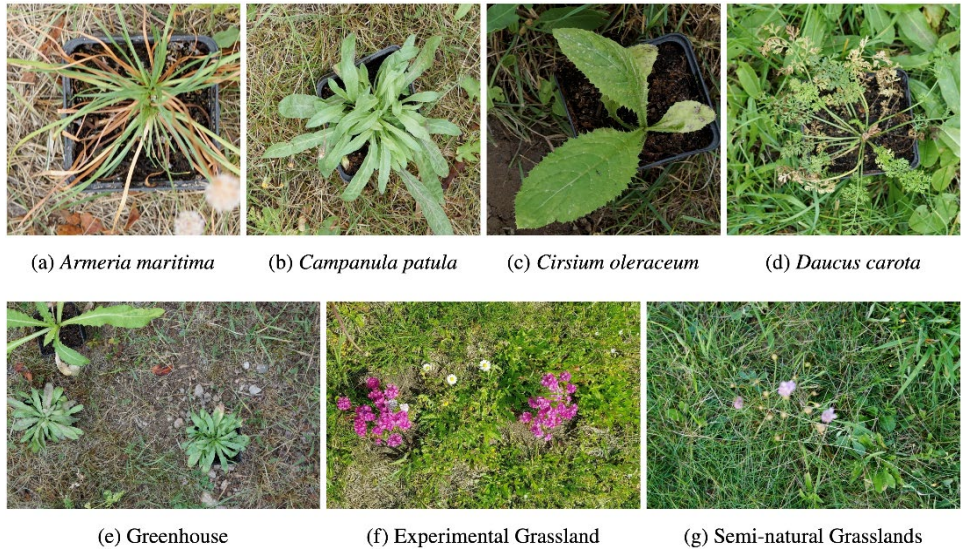

(a) *Armeria maritima*    (b) *Campanula patula*    (c) *Cirsium oleraceum*    (d) *Daucus carota*

(e) Greenhouse    (f) Experimental Grassland    (g) Semi-natural Grasslands

Fig. 2: The dataset consists of four indicator species (a-d) from three different sources (e-g).

## 3   Experiments

In this section, we present our experimental methodology and findings. We begin with an overview of the dataset and the experimental configuration in Section 3.1, followed by the experimental results in Section 3.2. Our experiments employ the accuracy metric known as *Average Precision* (AP), along with two localization metrics in XAI: *Attribution Localization* (AL) and *Top-K Intersection* (TKI). Detailed mathematical definitions for these metrics can be found in Appendix A.1.

### 3.1 Data and Modeling Setup

The dataset employed in our experiments was proposed by Basavegowda et al. [Ba22] and their subsequent collection endeavors. This dataset consists of RGB images captured from three different sources: greenhouse, an experimental grassland, and semi-natural grasslands, featuring four indicator species: *Armeria maritima*, *Campanula patula*, *Cirsium oleraceum*, and *Daucus carota*. An overview of this dataset is presented in Figure 2.

|                | Semi-natural Grassland | Experimental Grassland | Greenhouse |
|----------------|------------------------|------------------------|------------|
| Training Set   | 2485                   | 2440                   | 7515       |
| Validation Set | 529                    | 388                    | 1687       |
| Test Set       | 514                    | 401                    | 2043       |
| All            | 3528                   | 3229                   | 11245      |

Tab. 1: Dataset Partition for Training, Validation, and Testing

In our experiments, the dataset was randomly divided into three subsets: a training set, a validation set, and a test set, as shown in Table 1. This partitioning followed the common distribution practice, allocating roughly 60% for the training set, 20% for the validation set, and another 20% for the test set. We employed a range of data augmentation techniques, including rotation, clipping, and flipping, which infuse the dataset with variations in image viewpoints and orientations, enhancing its diversity.

All experiments were executed using 4 NVIDIA Tesla V100 GPUs over 200 training epochs with a batch size of 16. We employed an AdamW [LH19] optimizer with an initiate learning rate of 1e-3, complemented by a learning rate plateau scheduler with a patient threshold of 5. The pretrained weights used in the experiments were initially trained on the COCO dataset.

### 3.2 Results

**Transfer Learning.** We trained our detection model EfficientDet-D2 with different transfer learning settings: (i) without leveraging pretrained weights; (ii) using pretrained weights with a frozen feature extraction network; and (iii) utilizing pretrained weights with fine-tuning. All these experiments were trained with the combination of three training sets and were uniformly tested on the semi-natural test set.

The experimental evaluation in Table 2 revealed that fine-tuning the model with pretrained weights and without freezing any layers achieved the highest AP of 66.4, AL of 0.828, and TKI of 0.899. Notably, this setting exhibited a remarkable 28% improvement in AP, providing valuable insights for agricultural applications, specifically emphasizing the significance of incorporating pretrained weights and fine-tuning on the agricultural target dataset.

| Transfer Learning Method | Training / Validation Set | Test Set | AP ↑ | AL ↑ | TKI ↑ |
|---|---|---|---|---|---|
| Freezing pretrained weights | sNat+Expt+Green | sNat | 38.3 | 0.347±0.09 | 0.344±0.21 |
| None pretrained weights | sNat+Expt+Green | sNat | 57.3 | 0.713±0.10 | 0.690±0.21 |
| **Fine-tuning pretrained weight** | sNat+Expt+Green | sNat | **66.4** | **0.828±0.05** | **0.899±0.08** |

Tab. 2: Evaluation results of experiments with different transfer learning methods, tested on the semi-natural grassland test set. Fine-tuning pretrained models achieved the best performance in AP, AL and TKI metrics. Higher values are preferred for all metrics, where ± indicates variance. In this table, semi-natural grassland is abbreviated as sNat, experimental grassland as Expt, and greenhouse data as Green.

**Explainable AI.** In our quest to demonstrate the application of XAI in the field of grassland monitoring, we conducted the following three different experiments, where for each experiment the EfficientDet model was trained on a different dataset: (i) exclusively semi-natural grassland data; (ii) a combination of semi-natural and experimental data; and (iii) an amalgamation of semi-natural, experimental, and greenhouse data. Our primary objective was to gain deeper insights into the decision-making processes of deep learning models in grassland species detection. To this end, we employed Gradient-weighted Class Activation Mapping (GradCAM) [Se17], a technique that reveals the most critical regions in input images influencing the model's prediction.

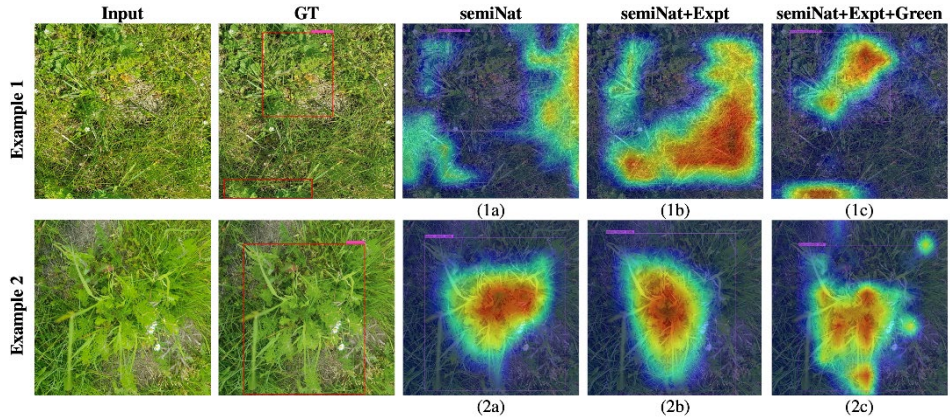

Fig. 3: First two columns represent input images and ground-truth annotations. Column 3-5 represent the predictions (purple bounding boxes) and explanations of experiments using different training data. In the explanation heat maps, the most relevant areas are highlighted in red and the less relevant areas in blue.

In Figure 3, we showcased two examples from the semi-natural grassland test set. The first example (row 1) clearly illustrated the prowess of the model trained with the entire

8      Shanghua Liu et al.

grassland training datasets. This experiment excels at emphasizing target features while minimizing focus on the background. In contrast, the model trained with semi-natural and experimental grassland data failed to detect the indicator plant, primarily due to significant focus on background, as shown in (1b). The second example (row 2) showed again the advantage of utilizing the entire grassland data, allowing the model to discern finer discriminative features, such as the unique leaf shape of indicator species. This is exemplified by the small circular area highlighted outside the predicted bounding box at the top right of (2c) in Figure 3.

In addition to qualitative visualizations, we quantitatively evaluated these experiments considering accuracy metrics such as *Average Precision* (AP) and localization metrics, specifically *Attribution Localization* (AL) and *Top-K Intersection* (TKI), as detailed in Table 3. The inclusion of experimental and greenhouse data alongside semi-natural grassland data led to performance improvements in all metrics, elevating AP from 61.8 to 66.4, AL from 0.71 to 0.828, and TKI from 0.697 to 0.899.

Our innovative incorporation of localization metrics into the evaluation framework expanded the spectrum of model evaluation. This extension allowed us not only to evaluate the detector's accuracy in identifying indicator species in the current test set, i.e. the semi-natural grassland scenario, but also to assess its adaptability in detecting indicator species in other areas characterized by diverse backgrounds, such as the wild natural grassland scenario.

| Training / Validation Set | Test Set | Transfer Learning | AP ↑ | AL ↑ | TKI ↑ |
|---|---|---|---|---|---|
| sNat | sNat | Fine-tuning | 61.8 | 0.710±0.14 | 0.697±0.21 |
| sNat+Expt | sNat | Fine-tuning | 63.6 | 0.735±0.16 | 0.764±0.18 |
| sNat+Expt+Green | sNat | Fine-tuning | **66.4** | **0.828±0.05** | **0.899±0.08** |

Tab. 3: Evaluation result of experiments with different training data, tested on the semi-natural grassland test set for fair comparison. In this table, semi-natural grassland is abbreviated as sNat, experimental grassland as Expt, and greenhouse data as Green. The experiment using all training sets achieved the best performance in AP, AL and TKI.

## 4    Conclusion

In this work, we addressed two main shortcomings of developing a deep learning model for grassland monitoring: the distributional shift between generalized and grassland-specific datasets and the inherent opacity of deep learning models. Our research highlights the remarkable potential of pretrained models, especially when combined with fine-tuning, to significantly boost model performance. By leveraging XAI, we gained a deeper understanding of the decision-making processes of deep learning models in the context of grassland scenarios, where features like the unique leaf shape of indicator species play an important role in decision-making. Furthermore, the utilization of localization metrics within XAI broadens the scope of model performance assessment, offering valuable insights into the model's generalizability and adaptability to other domains where the background is not necessarily class-informative. These findings hold significant implications for agricultural applications characterized by limited data and the need for domain adaptation. For future work, we will conduct experiments including additional XAI metrics and explore how their insights can incorporate model and dataset development in advancing AI-driven agricultural solutions.

# Appendix

### A.1    Metric

*Average Precision* (AP) is the crucial metric that offers a comprehensive evaluation of the model's object detection performance and is employed to determine the challenge winner in datasets like COCO. Calculating AP across all IoU thresholds and across all categories entails two steps:

$$AP = \frac{1}{N}\sum_{i=1}^{N} CAP_i(IoU_k) \tag{1}$$

Firstly, calculating class average precision at different IoU thresholds in equation (1), where $CAP_i$ represents the class average precision for class *i*, *N* is the number of IoU thresholds, and $CAP_i\ (IoU_k)$ denotes the class average precision for class *i* at threshold $IoU_k$. $CAP_i\ (IoU_k)$ value is calculated as the area under the precision-recall curve. We employ the same IoU threshold settings as COCO, ranging from 0.5 to 0.95 with a step size of 0.05 (i.e., *N* = 10).

$$AP = \frac{1}{C}\sum_{i=1}^{C} CAP_i \qquad (2)$$

Secondly, in equation (2), calculating the overall average precision (AP) by averaging the class average precision (CAP) over all different classes, where *AP* represents Average Precision metric and *C* stands for the total number of classes.

***Attribution Localization*** (AL) [Ko20] measures the ratio of positive attributions within the targeted object to the total positive attributions, computed as follows. Here, $R_{box}$ is the sum of positive relevance in the bounding box and $R_{tot}$ is the sum of positive relevance in the image.

$$AL = \frac{R_{box}}{R_{tot}} \qquad (3)$$

***Top-K Intersection*** (TKI) [TMBE22] computes the intersection between a bounding box and the binary explanation at the top k feature locations, where pixel-wise intersection is defined as *TKI* between the binary bounding box mask *M* and the binary mask of top-k features $E^k$. Here, *w* and *h* represents the width and height of the input image.

$$TKI = \frac{1}{k}\sum_{i=1}^{w}\sum_{j=1}^{h} M_{ij} \wedge E_{ij}^{k} \qquad (4)$$

We employed GradCAM [Se17] as an explanation tool for model predictions. This assessment involved measuring the alignment of features employed in the model predictions with the ground true position of the target (bounding box).